\documentclass[runningheads]{llncs}
\usepackage{comment}
\usepackage{graphicx}
\usepackage{amsmath}
\usepackage{cite}

\usepackage{color}
\usepackage{tikz}
\usepackage{graphicx}
\usepackage{caption}
\usepackage{subcaption}
\usepackage[normalem]{ulem}
\usepackage{multirow}
\usepackage{pdfpages}
\usepackage{xcolor}
\usepackage{hyperref}
\usepackage{float}
\usepackage{verbatim}
\usepackage{subcaption}

\hypersetup{
    colorlinks=true,
    linkcolor=blue,
    filecolor=magenta,      
    urlcolor=cyan,
}

\begin{document}
\title{Identification and Recognition of Rice Diseases and Pests Using Convolutional Neural Networks}
\titlerunning{Rice Disease and Pest Classification}
%
\author{Chowdhury Rafeed Rahman (UIU) \and
Preetom Saha Arko (BUET) \and
Mohammed Eunus Ali (BUET) \and
Mohammad Ashik Iqbal Khan (BRRI) \and
Sajid Hasan Apon (BUET) \and 
Farzana Nowrin (BRRI) \and 
Abu Wasif (BUET) \and 
}
\authorrunning{Chowdhury Rafeed Rahman}
%
\institute{United International University (UIU) \\
Bangladesh Rice Research Institute (BRRI) \\
Bangladesh University of Engineering and Technology (BUET)
}
\maketitle              
\begin{abstract}
An accurate and timely detection of diseases and pests in rice plants can help farmers in applying timely treatment on the plants and thereby can reduce the economic losses substantially. Recent developments in deep learning based convolutional neural networks (CNN) have greatly improved the image classification accuracy. Being motivated by the success of CNNs in image classification, deep learning based approaches have been developed in this paper for detecting diseases and pests from rice plant images. The contribution of this paper is two fold: (i) State-of-the-art large scale architectures such as VGG16 and InceptionV3 have been adopted and fine tuned for detecting and recognizing rice diseases and pests. Experimental results show the effectiveness of these models with real datasets. (ii) Since large scale architectures are not suitable for mobile devices, a two-stage small CNN architecture has been proposed, and compared with the state-of-the-art memory efficient CNN architectures such as MobileNet, NasNet Mobile and SqueezeNet. Experimental results show that the proposed architecture can achieve the desired accuracy of 93.3\%  with a significantly reduced model size (e.g., 99\% less size compared to that of VGG16).

\end{abstract}
\section{Introduction}
Rice occupies about 70 percent of the grossed crop area 
and accounts for 93 percent of total cereal production in Bangladesh \cite{coelli2002technical}. Rice also ensures food security of over half the world population \cite{fao}. Researchers have observed 10-15\% average yield loss because of 10 major diseases of rice in Bangladesh \cite{BRRI}. Timely detection of rice plant diseases and pests is one of the major challenges in agriculture sector. Hence, there is a need for automatic rice disease detection using readily available mobile devices in rural areas. 
\par 

Deep learning techniques have shown great promise in image classification. In recent years, these techniques have been used to analyse diseases of tea  \cite{karmokar2015tea}, apple \cite{wang2017automatic}, tomato \cite{fuentes2017robust}, grapevine, peach, and pear \cite{sladojevic2016deep}. \cite{babu2007leaves} proposed a feed forward back propagation neural network from scratch in order to detect the species of plant from leaf images. Neural network ensemble (NNE) was used by \cite{karmokar2015tea} to recognize five different diseases of tea plant from tea leaves. \cite{bhagawati2015artificial} trained a neural network with weather parameters such as temperature, relative humidity, rainfall and wind speed to forecast rice blast disease. \cite{mohanty2016using} used deep CNN to detect disease from leaves using 54306 images of 14 crop species representing 26 diseases, while \cite{sladojevic2016deep} used CaffeNet model to recognize 13 different types of plant diseases. \cite{wang2017automatic} worked on detecting four severity stages of apple black rot disease using \textit{PlantVillage} dataset. They used CNN architectures with different depths and implemented two different training methods on each of them. A real time tomato plant disease detector was built using deep learning by \cite{fuentes2017robust}. \cite{brahimi2017deep} used fine-tuned AlexNet and GoogleNet to detect nine diseases of tomatoes. \cite{cruz2017x} injected some texture and shape features to the fully connected layers placed after the convolutional layers so that the model can detect Olive Quick Decline Syndrome effectively from the limited dataset. Instead of resizing images to a smaller size and training a model end-to-end, \cite{dechant2017automated} used a three stage architecture (consisting of multiple CNNs) and trained the stage-one model on full scaled images by dividing a single image into many smaller images. \cite{barbedo2018impact} used transfer learning on GoogleNet to detect 56 diseases infecting 12 plant species. Using a dataset of 87848 images of leaves captured both in laboratory and in the field, \cite{ferentinos2018deep} worked with 58 classes containing 25 different plants. \cite{liu2018identification} built a CNN combining the ideas of AlexNet and GoogLeNet to detect four diseases of apple. Images of individual lesions and spots instead of image of whole leaf were used by \cite{barbedo2019plant} for identifying 79 diseases of 14 plant species. Few researches have also been conducted on rice disease classification~(\cite{lu2017identification,atole2018multiclass}). \cite{lu2017identification} conducted a study on detecting 10 different rice plant diseases using a small handmade CNN architecture, inspired by older deep learning frameworks such as LeNet-5 and AlexNet, using 500 images. \cite{atole2018multiclass} used AlexNet (large architecture) to distinguish among three classes - normal rice plant, diseased rice plant and snail infected rice plant using 227 images. 

Researches mentioned above mainly focused on accurate plant disease recognition and classification. For this purpose, they implemented various types of CNN architectures such as AlexNet, GoogLeNet, LeNet-5 and so on. In some studies, ensemble of multiple neural network architectures have been used. These studies played an important role for automatic and accurate recognition and classification of plant diseases. But their focus was not on modifying the training method for the models that they had constructed and used. Moreover, they did not consider the impact of the large number of parameters of these high performing CNN models in real life mobile application deployment.   

In this research, two state-of-the-art CNN architectures: VGG16 and InceptionV3 have been tested in various settings. Fine tuning, transfer learning and training from scratch have been implemented to assess their performance. In both the architectures, fine tuning the model while training has shown the best performance. Though these deep learning based architectures perform well in practice, a major limitation of these architectures is that they have large number of parameters, a problem similar to previously conducted researches. For example, there are about 138 million parameters in VGG16 \cite{VGG16}. In remote areas of developing countries, farmers do not have internet connectivity or have slow internet speed. So, a mobile application capable of running CNN based model offline is needed for rice disease and pest detection. So, a memory efficient CNN model with reasonably good classification accuracy is required. Since the reduction of the number of parameters in a CNN model reduces its learning capability, one needs to make a trade-off between memory requirement and classification accuracy to build such a model.

To address the above issue, in this paper, a new training method called \textbf{two stage training} has been proposed. A CNN architecture, namely \textbf{Simple CNN} has been proposed which achieves high accuracy leveraging two stage training in spite of its small number of parameters. Experimental study shows that the proposed \textbf{Simple CNN} model outperforms state-of-the-art memory efficient CNN architectures such as MobileNet, NasNet Mobile and SqueezeNet on recognizing rice plant diseases and pests.

All training and validation have been conducted on a rice dataset collected in real life scenario as part of this research. A rice disease may show different symptoms based on various weather and soil conditions. Similarly, pest attack can show different symptoms at different stages of an attack. Moreover, the diseases and pests can occur at any part of the plant which include leaf, stem and grain. Images can also be of heterogeneous background. This research addresses all these issues while collecting data. This paper focuses on recognizing eight different rice plant diseases and pests that occur at different times of the year at Bangladesh Rice Research Institute (BRRI). This work also includes a ninth class for non-diseased rice plant recognition.

In summary, this paper makes two important contributions in rice disease and pest detection. First, state-of-the-art large scale deep learning frameworks have been tested to investigate the effectiveness of these architectures in rice plant disease and pest identification from images collected from real-life environments. Second, a novel two-stage training based light-weight CNN has been proposed that is highly effective for mobile device based rice plant disease and pest detection. This can be an effective tool for farmers in remote environment.

\section{Materials and Methods}
\subsection{Data Collection}\label{Data}
Rice diseases and pests occur in different parts of the rice plant. Their occurrence depends on many factors such as temperature, humidity, rainfall, variety of rice plants, season, nutrition, etc. An extensive exercise was undertaken to collect total 1426 images of rice diseases and pests from paddy fields of Bangladesh Rice Research Institute (BRRI). Images have been collected in real life scenario with heterogeneous backgrounds from December, 2017 to June, 2018 for a total of seven months. The image collection has been performed in a range of weather conditions - in winter, in summer and in overcast condition in order to get as fully representative a set of images as possible. Four different types of camera have been used in capturing the images.
These steps increase the robustness of our model. This work encompasses total five classes of diseases, three classes of pests and one class of healthy plant and others - a total of nine classes. The class names along with the number of images collected for each class are shown in Table \ref{table:data collection}. It is to note that Sheath Blight, Sheath Rot and their simultaneous occurrence have been considered in the same class, because their treatment method and place of occurrence are the same.

\begin{table}[!h]
  \centering
	\begin{tabular}{|l|l|l|l|l|}
	\hline
	\textbf{Class Name} & \textbf{No. of Collected Images}\\ 
	\hline
	False Smut & 93\\
	\hline
	Brown Plant Hopper (BPH) & 71\\
	\hline
	Bacterial Leaf Blight (BLB) & 138\\
	\hline
	Neck Blast & 286\\
	\hline 
	Stemborer & 201\\
	\hline
	Hispa & 73\\
	\hline
	Sheath Blight and/or Sheath Rot & 219\\
	\hline
	Brown Spot & 111\\
	\hline
	Others & 234\\
	\hline
	
	\end{tabular}
	\caption{Image Collection of Different Classes}\label{table:data collection}
\end{table}

\begin{figure}[!h]
\captionsetup[subfigure]{labelformat=empty}
 \centering
 \begin{subfigure}{.30\textwidth}
  \centering
  \includegraphics[width=1\linewidth]{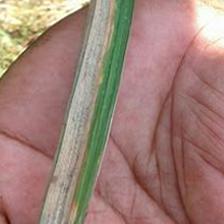}
  \caption{(a) Bacterial Leaf Blight (Disease)}
  \label{fig:BLB}
 \end{subfigure}\hfill%
 \begin{subfigure}{.30\textwidth}
  \centering
  \includegraphics[width=1\linewidth]{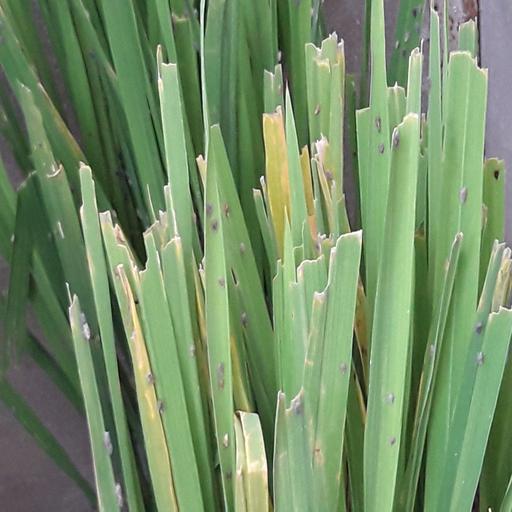}
  \caption{(b) Brown Plant Hopper (Pest)}
 \end{subfigure}\hfill%
 \begin{subfigure}{.30\textwidth}
  \centering
  \includegraphics[width=1\linewidth]{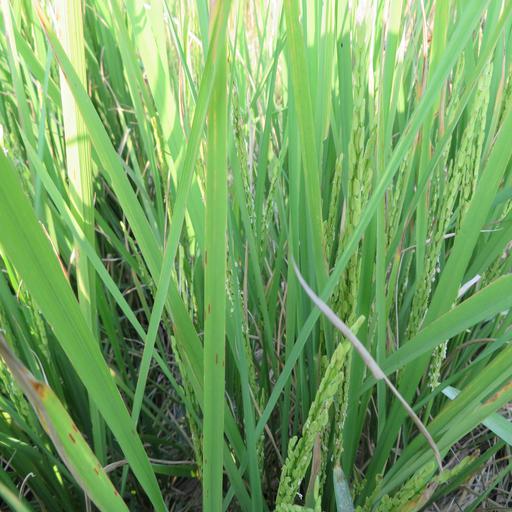}
  \caption{(c) Brown Spot (Disease)}
 \end{subfigure}
 
 \begin{subfigure}{.30\textwidth}
  \centering
  \includegraphics[width=1\linewidth]{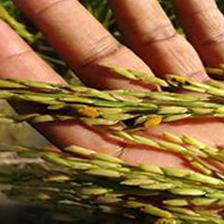}
  \caption{(d) False Smut (Disease)}
  \label{fig:FS}
 \end{subfigure}\hfill%
 \begin{subfigure}{.30\textwidth}
  \centering
  \includegraphics[width=1\linewidth]{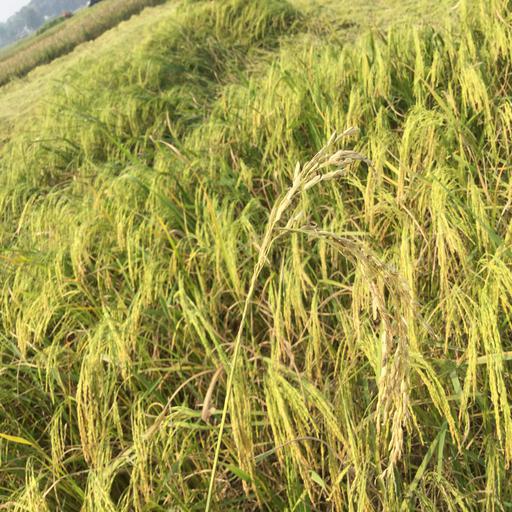}
  \caption{(e) Stemborer (Pest)}
 \end{subfigure}\hfill%
 \begin{subfigure}{.30\textwidth}
  \centering
  \includegraphics[width=1\linewidth]{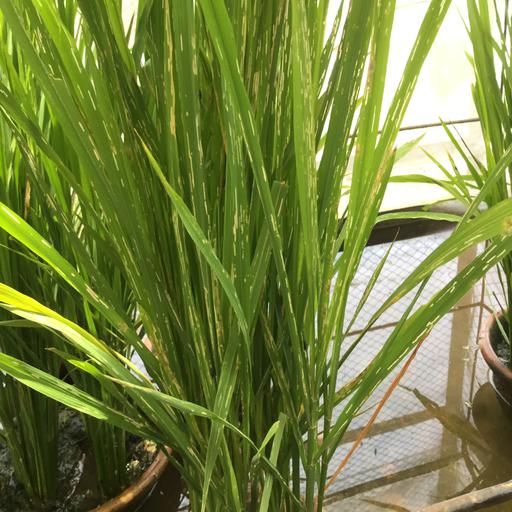}
  \caption{(f) Hispa (Pest)}
 \end{subfigure}

 \begin{subfigure}{.30\textwidth}
  \centering
  \includegraphics[width=1\linewidth]{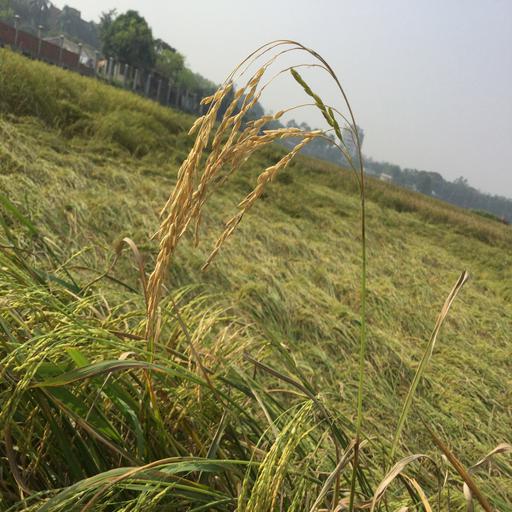}
  \caption{(g) Neck Blast (Disease)}
 \end{subfigure}\hfill%
 \begin{subfigure}{.30\textwidth}
  \centering
  \includegraphics[width=1\linewidth]{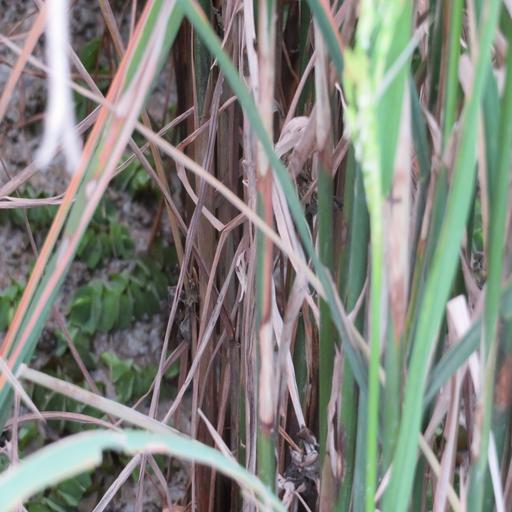}
  \caption{(h) Sheath Blight (Disease)}
 \end{subfigure}\hfill
 \begin{subfigure}{.30\textwidth}
  \centering
  \includegraphics[width=1\linewidth]{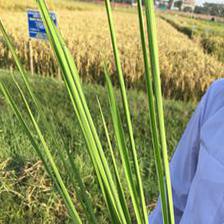}
  \caption{(i) Healthy Leaf (Others)}
  \label{fig:Oth}
 \end{subfigure}\hfill%
\caption{A Sample Image of Each Detected Class}
\label{fig:sample image of Each Disease and Pest}
\end{figure}

Symptoms of different diseases and pests are seen at different parts such as leaf, stem and grain of the rice plant. Bacterial Leaf Blight disease, Brown Spot disease, Brown Plant Hopper pest (late stage) and Hispa pest occur on rice leaf. Sheath Blight disease, Sheath Rot disease and Brown Plant Hopper pest (early stage) occur on rice stem. Neck Blast disease and False Smut disease occur on rice grain. Stemborer pest occurs on both rice stem and rice grain. All these aspects have been considered while capturing images. To prevent classification models from being confused between dead parts and diseased parts of rice plant, images of dead leaf, dead stem and dead grain of rice plants have been incorporated into the dataset. For example, diseases like BLB, Neck Blast and Sheath Blight have similarity with dead leaf, dead grain and dead stem of rice plant respectively. Thus images of dead leaf, dead stem and dead grain along with images of healthy rice plant have been considered in a class that has been named \textbf{others}. Sample images of each class have been depicted in Figure \ref{fig:sample image of Each Disease and Pest}. 

\begin{figure}[!h]
\captionsetup[subfigure]{labelformat=empty}
 \centering
 \begin{subfigure}{.30\textwidth}
  \centering
  \includegraphics[width=1\linewidth]{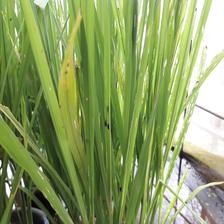}
  \caption{}
 \end{subfigure}\hfill%
 \begin{subfigure}{.30\textwidth}
  \centering
  \includegraphics[width=1\linewidth]{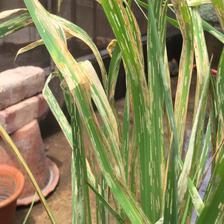}
  \caption{}
 \end{subfigure}
 \caption{\textbf{Hispa Variations:} Image on the left has visible black pests and white spots on plant leaf which occur during early stage of Hispa attack. Image on the right has intense spots on leaves with no visible pest occurring during later stage of Hispa attack}
 \label{fig: Hispa_var}
\end{figure}

False Smut, Stemborer, Healthy Plant class, Sheath Blight and/or Sheath Rot class show multiple types of symptoms. Early stage symptoms of Hispa and Brown Plant Hopper are different from their later stage symptoms. All symptom variations of these classes found in the paddy fields of BRRI have been covered in this work. These intra-class variations have been described in Table \ref{table:Symptom Variation}. BLB, Brown Spot and Neck Blast disease show no considerable intra-class variation around BRRI area. An illustrative example for Hispa pest has been given in Figure \ref{fig: Hispa_var}.  

\begin{table}[!h]
  \centering
	\begin{tabular}{|l|l|l|}
	\hline
	\textbf{Class Name} & \textbf{Symptom Variation} & \textbf{Sample No.}\\
	\hline
	\multirow{2}{*}{BPH} & Early stage of Brown Plant Hopper attack & 50\\
	                     \cline{2-3}
	                     & Late stage of Brown Plant Hopper attack  & 21\\
	\hline
	\multirow{2}{*}{False Smut} & Brown symptom & 66\\
	                     \cline{2-3}
	                     & Black symptom & 27\\
	\hline
	\multirow{2}{*}{Others} & Healthy green leaf and stem & 96\\
	                     \cline{2-3}
	                     & Healthy yellow grain & 71\\
	                     \cline{2-3}
	                     & Dead leaf and stem & 67\\
	\hline
	\multirow{2}{*}{Hispa} & Visible black pests and white spot on plant leaf & 53\\
	                     \cline{2-3}
	                     & No visible pest, intense spot on leaves & 20\\
	\hline 
	\multirow{2}{*}{Stemborer} & Symptom on grain & 180\\
	                     \cline{2-3}
	                     & Symptom on stem & 21\\
	\hline
	\multirow{2}{*}{Sheath Blight and/or Sheath Rot} & Black Stem & 70\\
	                     \cline{2-3}
	                     & White spots & 77\\
	                     \cline{2-3}
	                     & black and white symptom mixed & 72\\
	\hline
	\end{tabular}
	\caption{Intra-class Variation in Some Diseases and Pests}\label{table:Symptom Variation}
\end{table}

\subsection{Experimental Setup}
\href{https://keras.io/}{\textbf{Keras framework}} with tensorflow back-end has been used to train the models. Experiments have been conducted with two state-of-the-art CNN architectures containing large number of parameters such as \textbf{VGG16} and \textbf{InceptionV3}. Later the proposed light-weight two-stage \textbf{Simple CNN} have been tested and compared with three state-of-the-art memory efficient CNN architectures such as MobileNetv2, NasNet Mobile and SqueezeNet. VGG16 \cite{VGG16} is a sequential CNN architecture using 3$\times$3 convolution filters. After each maxpooling layer, the number of convolution filters gets doubled in VGG16. InceptionV3 \cite{Inception} is a non-sequential CNN architecture consisted of inception blocks. In each inception block, convolution filters of various dimensions and pooling are used on the input in parallel. The number of parameters of these five architectures along with \textbf{simple CNN} architecture have been given in Table \ref{table:Parameter}. Three different types of training methods have been implemented on each of these five architectures.

\begin{table}[!h]
  \centering
	\begin{tabular}{|l|l|}
	\hline
	\textbf{CNN Architecture} & \textbf{No. of Parameters}\\ 
	\hline
	VGG16 & 138 million\\
	\hline
	InceptionV3 & 23.8 million\\
	\hline
	MobileNetv2 & 2.3 million\\
	\hline
	NasNet Mobile & 4.3 million\\
	\hline 
	SqueezeNet & 0.7 million\\
	\hline
	Simple CNN & 0.8 million\\
	\hline
	\end{tabular}
	\caption{State-of-the-art CNN Architectures and Their Parameter No.}\label{table:Parameter}
\end{table}

\textbf{Baseline training:} All randomly initialized architecture layers are trained from scratch. This method of training takes time to converge.\par
\textbf{Fine Tuning: } The convolution layers of the CNN architectures are trained from their pre-trained ImageNet weights, while the dense layers are trained from randomly initialized weights.\par
\textbf{Transfer Learning:} In this method, the convolution layers of the CNN architectures are not trained at all. Rather pre-trained ImageNet weights are kept intact. Only the dense layers are trained from their randomly initialized weights.\par

\textbf{10-fold cross-validation accuracy} along with \textbf{standard deviation} have been used as model performance metric since the dataset used in this work does not have any major imbalance. \textbf{Categorical Crossentropy} has been used as loss function for all CNN architectures since this work deals with multi-class classification. All intermediate layers of the CNN architectures used in this work have \textbf{relu} as activation function while the activation function used in the last layer is \textbf{softmax}. The hyperparameters used are as follows: dropout rate of 0.3, learning rate of 0.0001, mini batch size of 64 and number of epochs 100. These values have been obtained through hyperparamter tuning using 10-fold cross-validation. Adaptive Moment Estimation (Adam) optimizer has been used for updating the model weights.\par

All the images have been resized to the default image size of each architecture before working with that architecture. For example, InceptionV3 requires 299$\times$299$\times$3 pixel size image while VGG16 requires image of pixel size 224$\times$224$\times$3. Random rotation from -15 degree to 15 degree, rotations of multiple of 90 degree at random, random distortion, shear transformation, vertical flip, horizontal flip, skewing and intensity transformation have been used as part of the data augmentation process. Every augmented image is the result of a particular subset of all these 
transformations, where rotation type transformations have been assigned high probability. It is because CNN models in general are not rotation invariant. In this way, 10 augmented images from every original image have been created. Random choice of the subset of the transformations helps augment an original image in a heterogeneous way. 

A remote Red Hat Enterprise Linux server of RMIT University has been used for carrying out the experiments. The configuration of the server includes 56 CPUs, 503 GB RAM, 1 petabyte of user specific storage and two NVIDIA Tesla P100-PCIE GPUs each of 16 GB.

\subsection{Proposed \textbf{Simple CNN} Model}
\begin{figure}[!h]
\captionsetup[subfigure]{labelformat=empty}
 \centering
 \begin{subfigure}{0.45\textwidth}
  \centering
  \includegraphics[width=1\linewidth]{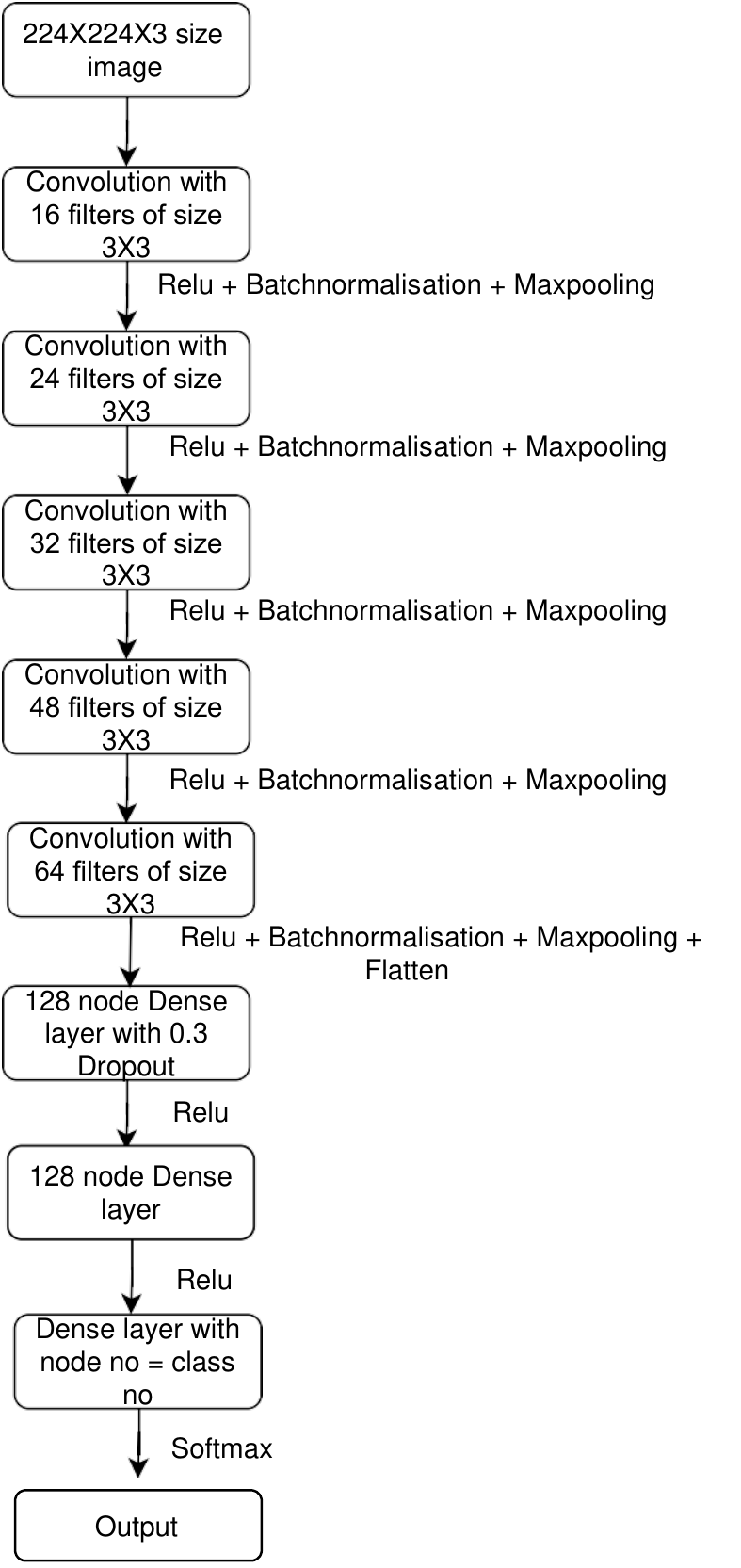}
 \end{subfigure}
\caption{Simple CNN Architecture}\label{fig: Hand_CNN}
\end{figure}

Apart from adapting state-of-the-art CNN models, a memory efficient two-stage small CNN architecture, namely \textbf{Simple CNN} shown in Figure  \ref{fig: Hand_CNN} has been constructed from scratch inspired by the sequential nature of VGG16. Fine tuned VGG16 provides excellent result on rice dataset. This \textbf{Simple CNN} architecture has only 0.8 million parameters compared to 138 million parameters of VGG16. All five of the state-of-the-art CNN architectures trained and tested in this work have shown the best result when fine tuning has been used (see Section \ref{Evaluation}). Two stage training is inspired from fine tuning. In \textbf{stage one}, the entire image dataset of nine classes are divided into 17 classes by keeping all intra-class variations in separate classes. These variations have been shown in detail in Table \ref{table:Symptom Variation}. For example, \textbf{others} class is divided into three separate classes. Thus, the model is trained with this 17 class dataset. As a result, the final dense layer of the model has 17 nodes with softmax activation function.
In \textbf{stage two}, the original dataset of nine classes is used. All layer weights of simple CNN architecture obtained from stage one are kept intact except for the topmost layer. This dense layer consisting of 17 nodes is replaced with a dense layer consisting of nine nodes with softmax activation function. Such measures are taken, because stage two training data are divided into the nine original classes. Now all the layers of the \textbf{Simple CNN} architecture are trained using this nine class dataset which are initialized with the pre-trained weights obtained from stage one training. Experiments show the effectiveness of applying this method.

\section{Results and Discussion}\label{Evaluation}
Experimental results obtained from 10-fold cross-validation for the five state-of-the-art CNN architectures along with Simple CNN have been shown in Table \ref{table:results}.  Transfer learning gives the worst result in all five of the models. For the smallest architecture SqueezeNet, it is below 50\%. Rice disease and pest images are different from images of ImageNet dataset. Hence, the freezing of convolution layer weights disrupts learning of the CNN architectures. Although baseline training also known as training from scratch does better than transfer learning, the results are still not satisfactory. For the three small models, the accuracy is less than 80\%. The standard deviation of validation accuracy is also large which denotes low precision. This shows that the models are not being able to learn the distinguishing features of the classes when trained from randomly initialized weights. More training data may solve this problem. Fine tuning gives the best result in all cases. It also ensures high precision (lowest standard deviation). It means that for the state-of-the-art CNN architectures to achieve good accuracy on rice dataset, training on the large ImageNet dataset is necessary prior to training on the rice dataset. Fine-tuned VGG16 achieves the best accuracy of 97.12\%. The \textbf{Simple CNN} architecture utilizing two stage training achieves comparable accuracy and the highest precision without any prior training on ImageNet dataset. Rather, this model is trained from scratch. 

From Table \ref{table:Parameter}, it is evident that the Simple CNN model has small number of parameters even when compared to small state-of-the-art CNN architectures such as MobileNet and NasNet Mobile. The number of parameters of SqueezeNet (the smallest of the five state-of-the-art CNN architectures in terms of parameter number) is comparable to the parameter number of the Simple CNN. Sequential models like VGG16 need depth in order to achieve good performance, hence have large number of parameters. Although Simple CNN is a sequential model with low number of parameters, its high accuracy (comparable to the other state-of-the-art CNN architectures) proves the effectiveness of two stage training. Future research should aim at building miniature version of memory efficient non-sequential state-of-the-art CNN architectures such as InceptionV3, DenseNet and Xception. These architectures should be able to achieve similar excellent result with even smaller number of parameters. 

A major limitation of two stage training is that the entire dataset has to be divided manually into symptom classes. In a large dataset, detecting all the major intra-class variations is a labour intensive process. There is a great chance of missing some symptom variations. Minor variety within a particular class maybe misinterpreted as separate symptom. One possible solution could be to use high dimensional clustering algorithms on each class specific image set separately in order to automate this process of identifying intra-class variations.    

\begin{figure}[!h]
 \centering
  \includegraphics[width=0.9\linewidth]{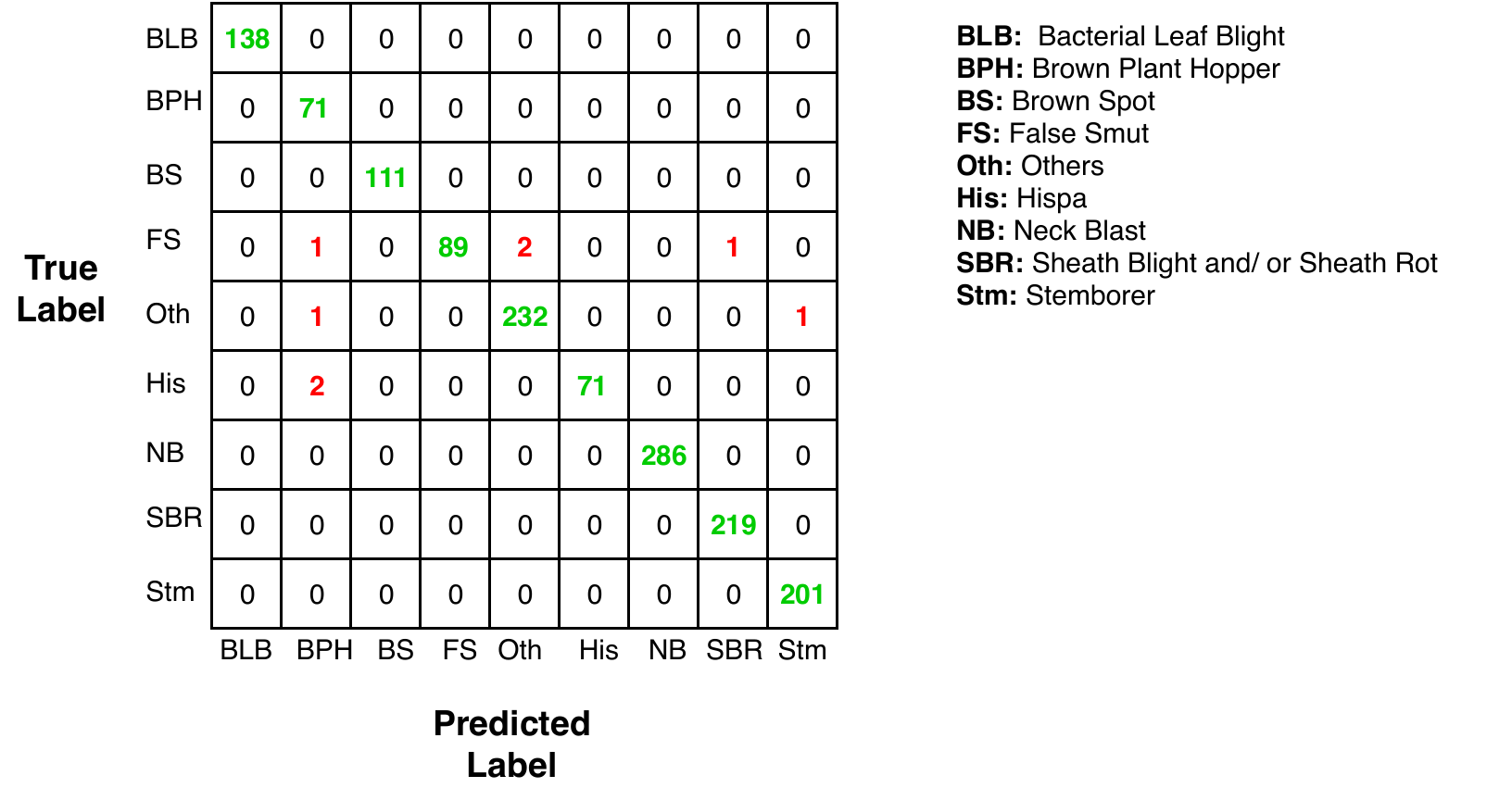}
 \caption{Confusion Matrix Generated Using Simple CNN on Entire Dataset}
 \label{fig:confution}
\end{figure}

 The confusion matrix generated from the application of Simple CNN on the entire dataset (training and validation set combined) has been shown in Figure \ref{fig:confution}. 4.3\% of the False Smut images existing in the dataset have been misclassified, which is the highest among all present classes of this work. False Smut symptom covers small portion of the entire image (captured in heterogeneous background) compared to other existing pest and disease images.    
 
\begin{figure}[!h]
 \centering
  \includegraphics[width=0.9\linewidth]{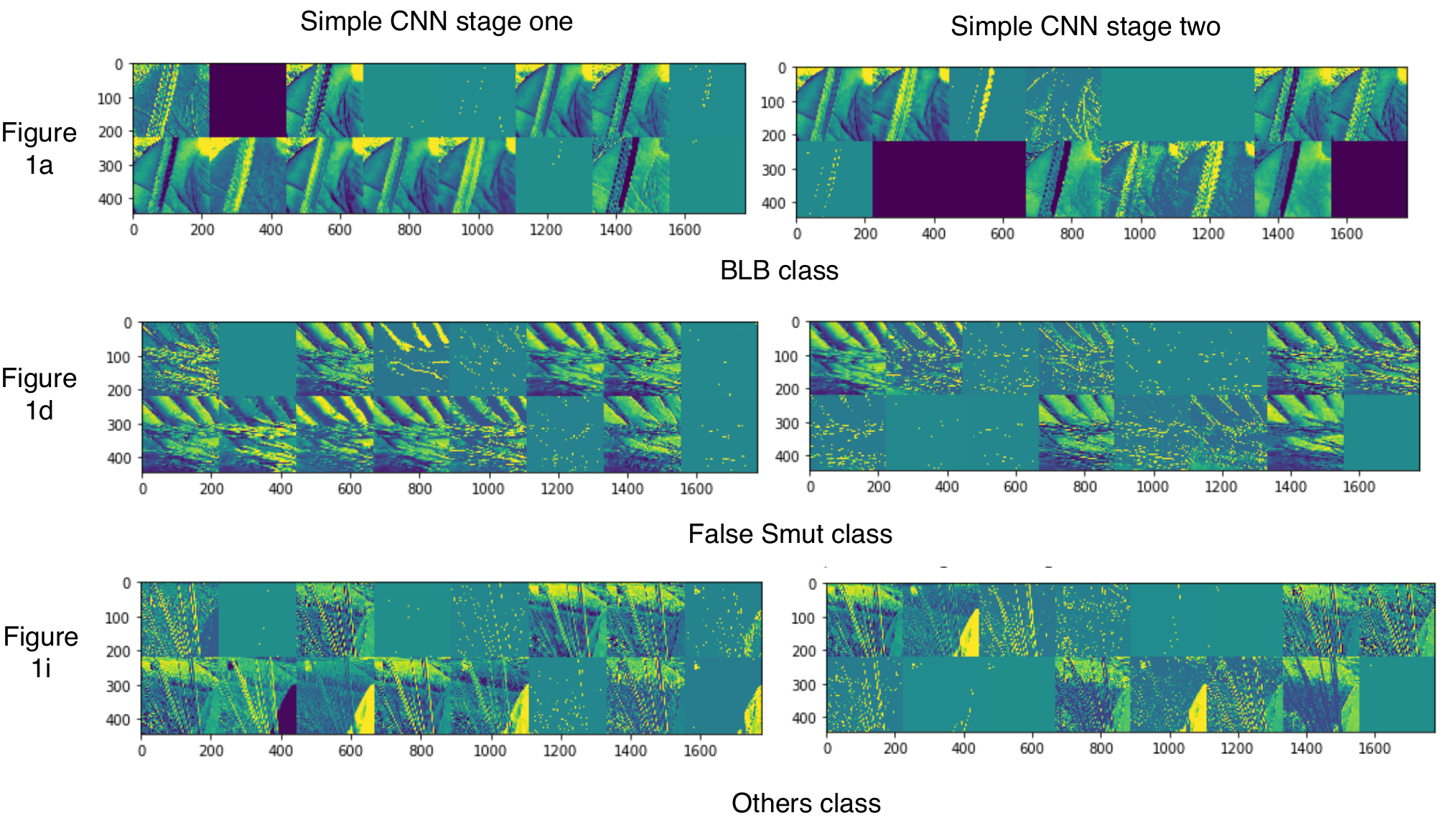}
 \caption{First Convolution Layer Output of Simple CNN}
 \label{fig:First_Inter_layer}
\end{figure}

\begin{figure}[!h]
 \centering
  \includegraphics[width=0.7\linewidth]{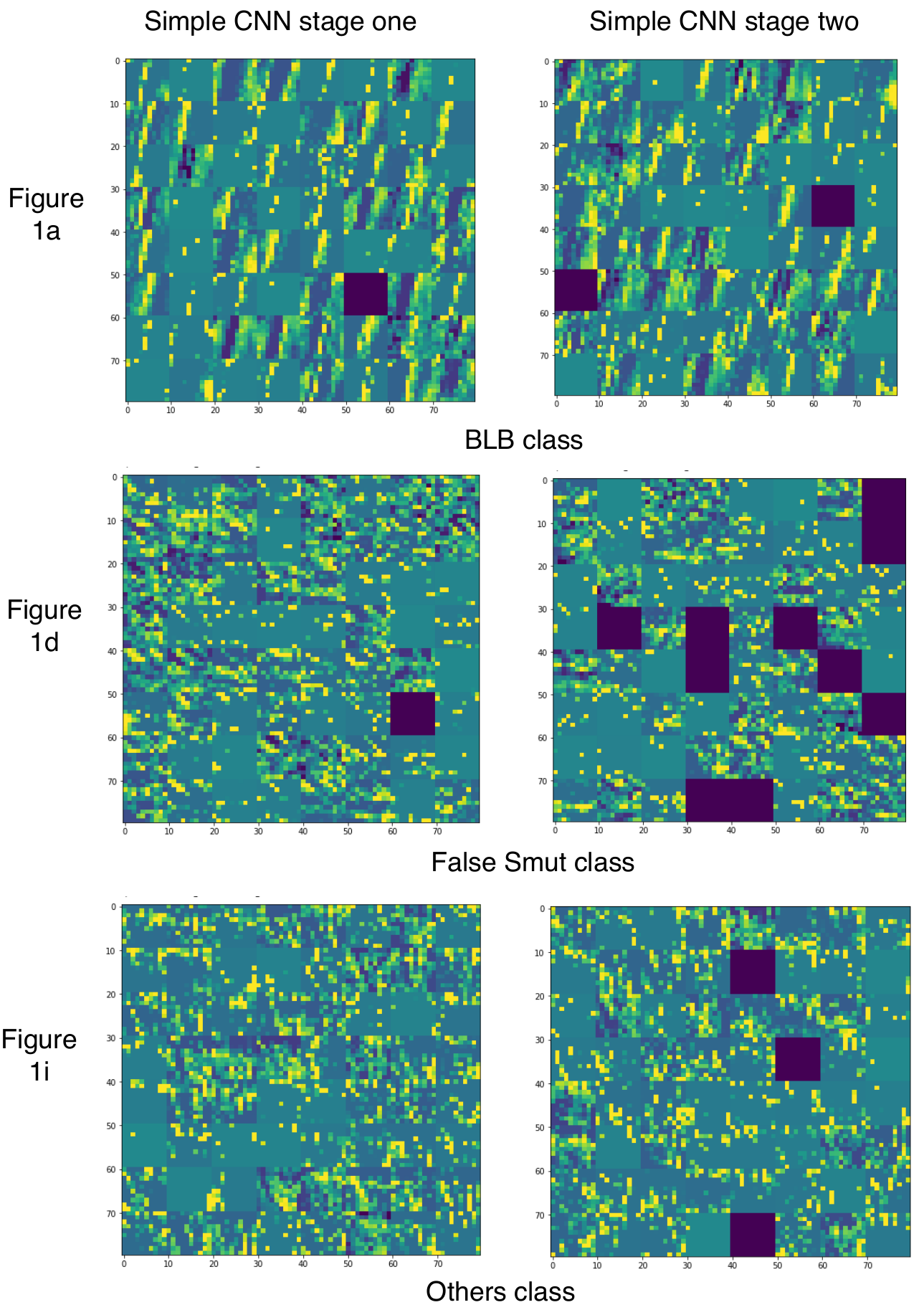}
 \caption{Last Convolution Layer Output of Simple CNN}
 \label{fig:Last_Inter_layer}
\end{figure}

 The first convolution layer outputs of Simple CNN have been shown in Figure \ref{fig:First_Inter_layer}. The three rows from top to bottom represent output for Figure \ref{fig:BLB}, Figure \ref{fig:FS} and Figure \ref{fig:Oth} respectively, while the left and right column represent output for stage one and stage two of the Simple CNN model respectively. Each of the six images contains 16 two dimensional mini images of size $222\times222$ (first convolution layer outputs a matrix of size $222\times222\times16$). The last convolution layer outputs of Simple CNN has been shown in Figure \ref{fig:Last_Inter_layer} in a similar setting. Each of the six images of Figure \ref{fig:Last_Inter_layer} contains 64 two dimensional mini images of size $10\times10$ (last convolution layer outputs a matrix of size $10\times10\times64$). The first layer maintains the regional features of the input image, although some of the filters are blank (not activated). The activations retain almost all of the information present in the input image. The last convolution layer outputs are visually less understandable. This representation depicts less information about the visual contents of the input image. Rather this layer attempts in presenting information related to the class of the image. The intermediate outputs for different classes are visually different for different classes. An interesting aspect can be observed in Figure \ref{fig:Last_Inter_layer}. Last convolution layer output for stage one model carries considerably less number of blank two dimensional mini images than does stage two model. This shows the capability of stage two model in terms of learning with less features. This helps Simple CNN achieve good accuracy and high precision after stage two training.

\begin{table}[]
\begin{tabular}{|c|c|c|c|}
\hline
\textbf{\begin{tabular}[c]{@{}c@{}}CNN \\ Architecture\end{tabular}} & \textbf{\begin{tabular}[c]{@{}c@{}}Training Method\\  Used\end{tabular}} & \textbf{\begin{tabular}[c]{@{}c@{}}Mean Validation\\  Accuracy\end{tabular}} & \textbf{\begin{tabular}[c]{@{}c@{}}Standard \\ Deviation\end{tabular}} \\ \hline
\multirow{3}{*}{VGG16}                                               & Baseline training                                                        & 89.19\%                                                                      & 10.28                                                                  \\ \cline{2-4} 
                                                                     & Transfer Learning                                                        & 86.52\%                                                                      & 5.37                                                                   \\ \cline{2-4} 
                                                                     & Fine Tuning                                                              & \textbf{97.12\%}                                                             & 2.23                                                                   \\ \hline
\multirow{3}{*}{InceptionV3}                                         & Baseline training                                                        & 91.17\%                                                                      & 3.96                                                                   \\ \cline{2-4} 
                                                                     & Transfer Learning                                                        & 72.09\%                                                                      & 7.96                                                                   \\ \cline{2-4} 
                                                                     & Fine Tuning                                                              & \textbf{96.37\%}                                                             & 3.9                                                                    \\ \hline
\multirow{3}{*}{MobileNetv2}                                         & Baseline training                                                        & 78.84\%                                                                      & 7.38                                                                   \\ \cline{2-4} 
                                                                     & Transfer Learning                                                        & 77.52\%                                                                      & 8.56                                                                   \\ \cline{2-4} 
                                                                     & Fine Tuning                                                              & \textbf{96.12\%}                                                             & 3.08                                                                   \\ \hline
\multirow{3}{*}{NasNet Mobile}                                       & Baseline training                                                        & 79.98\%                                                                      & 6.96                                                                   \\ \cline{2-4} 
                                                                     & Transfer Learning                                                        & 78.21\%                                                                      & 8.09                                                                   \\ \cline{2-4} 
                                                                     & Fine Tuning                                                              & \textbf{96.95\%}                                                             & 3.35                                                                   \\ \hline
\multirow{3}{*}{SqueezeNet v1.1}                                     & Baseline training                                                        & 74.88\%                                                                      & 8.18                                                                   \\ \cline{2-4} 
                                                                     & Transfer Learning                                                        & 42.76\%                                                                      & 9.12                                                                   \\ \cline{2-4} 
                                                                     & Fine Tuning                                                              & \textbf{92.5\%}                                                              & 3.75                                                                   \\ \hline
Simple CNN                                                          & Two Stage Training                                                       & \textbf{94.33\%}                                                             & 0.96                                                                   \\ \hline
\end{tabular}
\caption[]{Quantitative Performance of Different State-of-the-art CNN Architectures Obtained from 10 Fold Cross Validation\footnotemark}
\label{table:results}
\end{table}
\footnotetext{Best accuracy of each architecture has been mentioned in bold character.}

\section{Conclusion}\label{Conclusion}
This work has the following contributions:
\begin{itemize}
\item A dataset of rice diseases and pests consisting of 1426 images has been collected in real life scenario which cover eight classes of rice disease and pest. This dataset is expected to facilitate further research on rice diseases and pests. The dataset is available in the following link:
\url{https://drive.google.com/open?id=1ewBesJcguriVTX8sRJseCDbXAF_T4akK}.
The details of the dataset have been described in Subsection \ref{Data}.
\item Three different training methods have been implemented on two state-of-the-art large CNN architectures and three state-of-the-art small CNN architectures (targeted towards mobile applications) on rice dataset. Fine tuning from pre-trained ImageNet weight always provided the best result for all five architectures.
\item A new concept of two stage training derived from the concept of fine tuning has been introduced which enables proposed \textbf{Simple CNN} architecture of this work to perform well in real life scenario.
\end{itemize}

In future, incorporating location, weather and soil data along with the image of the diseased part of the plant can be investigated for a comprehensive and automated plant disease detection mechanism. Segmentation or object detection based algorithm can be implemented with a view to detecting and classifying rice diseases and pests more effectively in the presence of heterogeneous background.    



\section{Acknowledgments}
We thank Bangladesh ICT division for funding this research.
We also thank the authority of Bangladesh Rice Research Institute for supporting the research by providing us with the 
opportunity of collecting images of rice plant diseases in real life scenario. We also acknowledge the help of RMIT University who gave us 
the opportunity to use their GPU server. The authors would like to thank anonymous reviewers for providing suggestions which improved the quality of this article.

\bibliographystyle{splncs04}

\bibliography{main}

\end{document}